\documentclass[a4paper, conference]{IEEEtran}
\IEEEoverridecommandlockouts
\usepackage{cite}
\usepackage{amsmath,amssymb,amsfonts}
\usepackage{algorithm, algorithmic}
\usepackage{graphicx}
\usepackage{textcomp}
\usepackage{xcolor}
\usepackage{subfigure}
\usepackage{multirow}
\usepackage{makecell}
\usepackage{bm}

\begin{document}
	
	\title{Efficient Motion Planning for Automated Lane Change based on Imitation Learning and Mixed-Integer Optimization\\
		\thanks{*Indicate equal contribution. **corresponding author}
	}
	
	\author{\IEEEauthorblockN{Chenyang Xi*}
		\IEEEauthorblockA{\textit{School of Aerospace Engineering} \\
			\textit{Beijing Institute of Technology}\\
			Beijing, China \\
			caralhsi@gmail.com}
		\and
		\IEEEauthorblockN{Tianyu Shi*}
		\IEEEauthorblockA{\textit{Department of Civil Engineering} \\
			\textit{McGill University}\\
			Montreal, Canada \\
			tianyu.shi3@mail.mcgill.ca}
		\and
		\IEEEauthorblockN{ Yuankai Wu}
		\IEEEauthorblockA{\textit{Department of Civil Engineering} \\
			\textit{McGill University}\\
			Montreal, Canada \\
			yuankai.wu@mcgill.ca}
		\and
		\IEEEauthorblockN{\centerline{Lijun Sun**}}
		\IEEEauthorblockA{\textit{Department of Civil Engineering} \\
			\textit{McGill University}\\
			Montreal, Canada \\
			lijun.sun@mcgill.ca}
	}
	
	\maketitle
	
	\begin{abstract}
		
		Intelligent motion planning is one of the core components in automated vehicles, which has received extensive interests. Traditional motion planning methods suffer from several drawbacks in terms of optimality, efficiency and generalization capability. Sampling based methods cannot guarantee the optimality of the generated trajectories. Whereas the optimization-based methods are not able to perform motion planning in real-time, and limited  by the simplified formalization. In this work, we propose a learning-based approach to handle those shortcomings. Mixed Integer Quadratic Problem based optimization (MIQP) is used to generate the optimal lane-change trajectories which served as the training dataset for learning-based action generation algorithms. A hierarchical supervised learning model is devised to make the fast lane-change decision. Numerous experiments have been conducted to evaluate the optimality, efficiency, and generalization capability of the proposed approach. The experimental results indicate that the proposed model outperforms several commonly used motion planning baselines.
	\end{abstract}
	
	
	\section{Introduction}
	Automated vehicles (AVs) have recently become a popular subject in both research and industry domains, it is believed that AVs might bring about the societal and industrial revolution. AVs employ a wide range of technologies including the visual lidar perception, the sensor fusion, high-definition map (HD map) and localization, the prediction, the decision making, the motion planning and motion control, etc \cite{badue2019self}. A key issue in guaranteeing the safety of AVs is the generation of trajectories based on possible trajectories of other traffic participants, which is highly corresponding to the motion planning technologies. The problem investigated in this work was restricted to motion planning in lane-change scenarios. When an automated vehicle performed a lane-change movement, both the leading vehicle in the current lane and surrounding vehicles in the target lane should be carefully considered to ensure safety and comfort. Sampling based methods \cite{kuwata2009real}, suffer from the issues of optimality and efficiency. Optimization based methods \cite{rasekhipour2016potential,ji2016path}, suffer from the issues of efficiency and less generalizable capability, due to the simplification of constraints when formulating the problem (e.g., \cite{MIQP}). 
	
	There also existed a large number of works investigating the motion planning for lane-change maneuver from the imitation learning perspective. Deep Neural Networks was introduced to generalize driving policies from training data set collected from either human expert or traffic flow simulation \cite{shi2019data,sun2017fast,mehta2018learning}. Ashish Mehta et. Al. \cite{mehta2018learning} proposed a Multi-task Learning from Demonstration (MT-LfD) framework, supervised auxiliary task prediction was used to guide the main task of predicting the driving commands, and an end-to-end trainable network for imitating the expert demonstrator’s driving commands was introduced. However, due to the inherent characteristic of end-to-end imitation learning, there was an inevitable cascading failure. Therefore, the learned driving policy might not generalize to various scenarios since most of the data were collected in situations with limited variations. Conversely, deep reinforcement learning could directly learn policies mapping state variables to control commands without expert demonstrators. Reference \cite{wang2018reinforcement} introduced a reinforcement learning based approach to train the automated vehicle agent to learn an automated lane-change behavior such that it can intelligently make a lane-change decision under diverse and even unforeseen scenarios. However, to successfully deploy the proposed approach to real-world systems, the inevitable gap between simulation and real-world environment should be well addressed.
	
	Combining machine learning with conventional motion planning methods has already demonstrated effectiveness and efficiency in the literature \cite{zhang2016learning}. In \cite{zhang2016learning}, the authors proposed a novel method, MPC-Guided Policy Search, enabling control of automated aerial vehicles in complex and unstructured environments. Two main disadvantages of MPC were addressed: computation efficiency and requirement of full observation of the system. Reference \cite{de2019learning} introduced an approach to learning a lattice planner control (LLPC) set, the controller was learned from a representative dataset of vehicle paths and shown to be suitable for one particular task. This model is able to imitate the driving style of the learning path and easy to compute. Despite its strong capabilities, the optimality of its planned trajectories is not guaranteed.

	In this work, we proposed a learning based approach which overcame the aforementioned shortcomings, and several active lane-change scenarios have been used as the benchmark experiments. MIQP was used to generate the optimal lane-change trajectories which served as the training dataset, and Support Vector Machine (SVM) and Multi-Layer Perception (MLP) were deployed to learn the decision making and motion planning modules, respectively. 
	By doing so, we guarantee that the learning datasets of the proposed model are nearly optimal. In addition, the execution speed of the proposed model is sustainable as it is equal to input-output mapping through neural networks.

	\section{Formulation of MIQP}\label{sec:MIQP}
\begin{figure}[htbp]
		\centerline{\includegraphics[height=3.00cm]{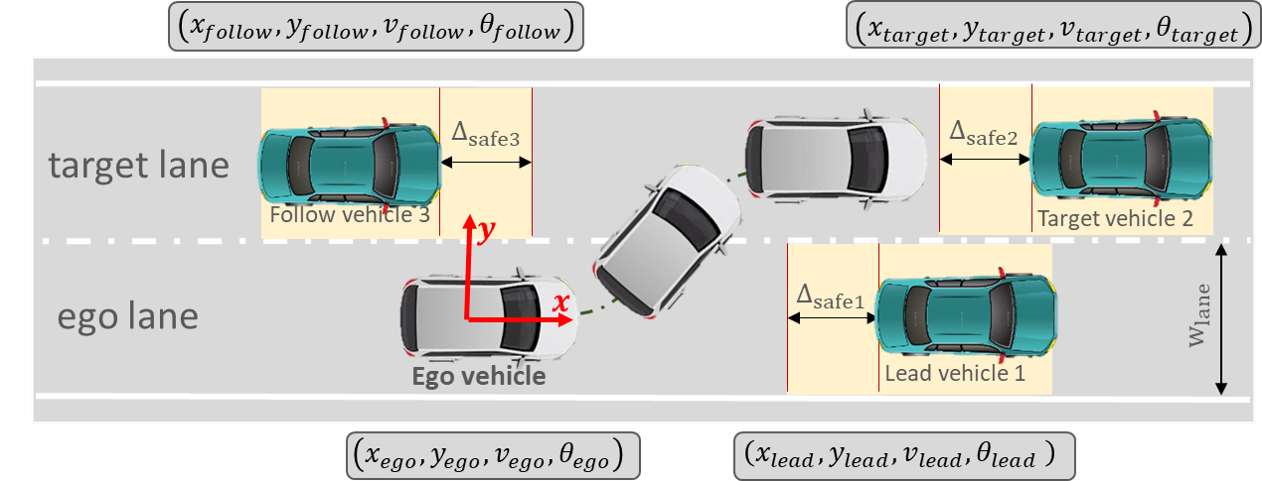}}
		\caption{lane-change scenario construction.}
		\label{lane_change_construction}
	\end{figure}

	 A typical lane-change scenario is displayed in Fig. \ref{lane_change_construction}. The ego vehicle (EV) was intended to take a lane-change maneuver to its left side. During the movement, three surrounding obstacle vehicles should be considered by the motion planning algorithm: the leading vehicle in the ego vehicles' current lane (ego lane, EL), and two nearest vehicles in the target lane (TL).

	Specifically, to satisfy the obstacle avoidance constraints and vehicle dynamics constraints, the ego vehicle needs to determine a suitable lateral yaw rate and longitudinal acceleration in a given time horizon, steering into the TL. In addition to ensuring the success of lane changing, the motion planning module should also guarantee the efficiency and safety of the ego vehicle’s movement (small acceleration and jerk). Also, the motion planning module should satisfy a safe distance between the EV and surrounding obstacle vehicles. 
	
	It is known that solving an optimization function that integrated with nonlinear constraints and cost function is a challenging job. In this section, we solve the lane-change problem by locally linearizing the dynamics equations and transferring the obstacle avoidance issue into logical constraints. Then, the problem is transformed into a MIQP  and can be solved by iterative optimization. Beyond this, a rational initialization of state variables is given for accelerating the optimization process.
	
	\subsection{Formulation of logical constraints}
	As shown in Fig.\ref{lane_change_construction}, in order to avoid collision with three obstacle vehicles, when the EV is in the EL, it should avoid Vehicle $1$. When the EV is in the TL, Vehicle $2$ and Vehicle $3$ should be avoided. We assume that the EV cannot exceed Vehicle $1$ in the EL. In addition, we consider the safety distance. The logical obstacle avoidance constraints are expressed by Eq. \ref{logic1}.
	
	\begin{equation}
	\left
	\{ 
	\begin{aligned} 
	x_{ego} &\leq x_{leader} - \Delta_{safe1}, \ \text{if}\  EV \in EL \\
	x_{ego} &\geq x_{follow} + \Delta_{safe2}, \ \text{if}\  EV \in TL \\
	x_{ego} &\leq x_{target} - \Delta_{safe3}, \ \text{if}\  EV \in TL \\
	\end{aligned} 
	\right. 
	\label{logic1}
	\end{equation}
	Where $x_{ego}$, $x_{leader}$, $x_{follow}$, and $x_{target}$ represent the longitudinal position of the EV and three surrounding vehicles respectively. $\Delta_{safe1}$, $\Delta_{safe2}$, and $\Delta_{safe3}$ represent the safe distances between the EV and the surrounding vehicles.
	
\begin{figure}[htbp]
		\centerline{\includegraphics[height=5.50cm]{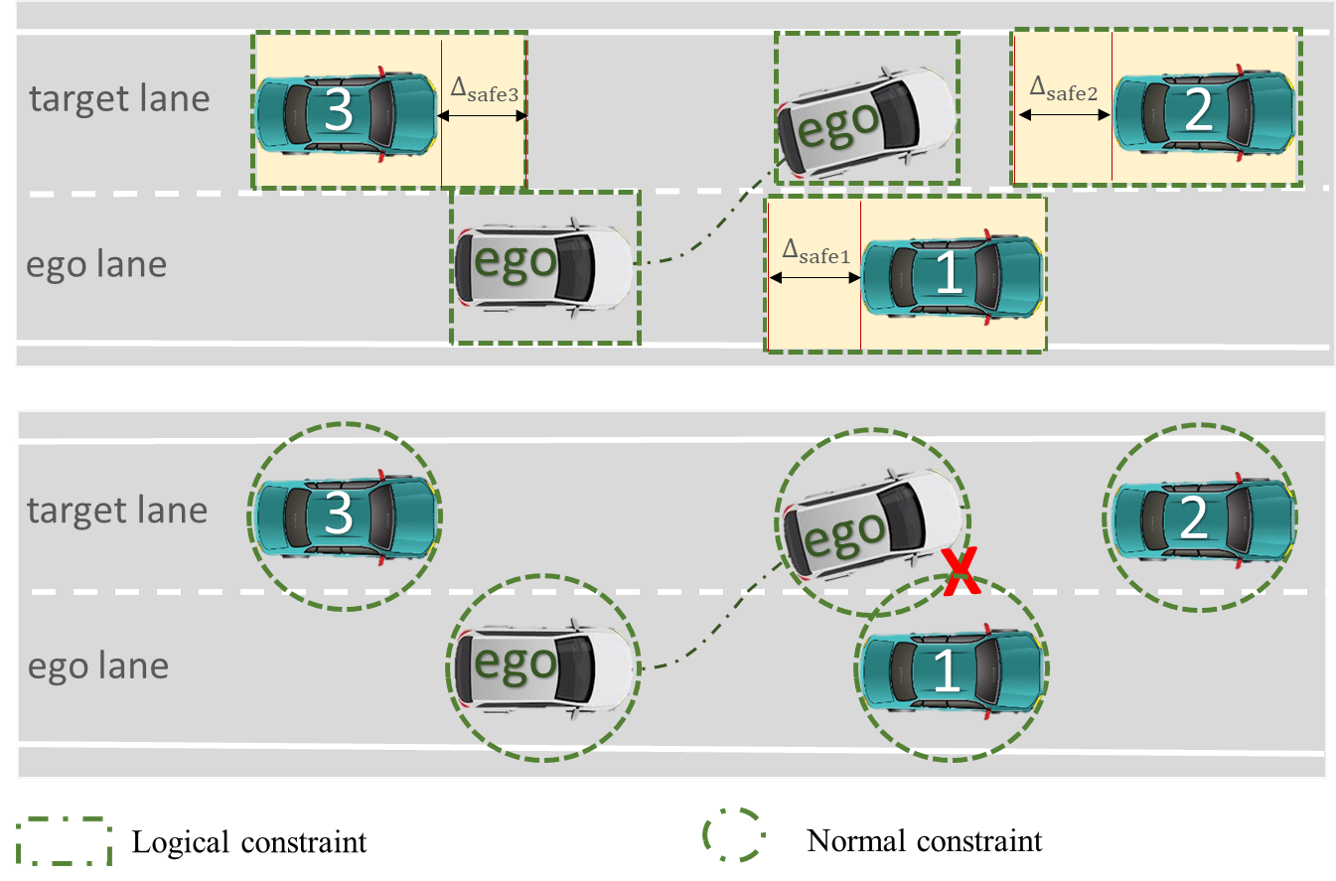}}
		\caption{Comparison of logical constraints and normal constraints}
		\label{Comparison of logical constraints and normal constraints}
	\end{figure}

	Logical constraints help us deal with the congested scenario, as shown in Fig. \ref{Comparison of logical constraints and normal constraints}. In this research \cite{sun2017fast}, surrounding vehicles are treated as circles. By doing so, such a congested scenario shown in Fig. \ref{Comparison of logical constraints and normal constraints} would not be solved, and the lane-change would not be processed. On the contrary, by using logical constraints, a feasible path can be conducted in such scenarios.
	
	\subsection{Optimization problem for lane-change}
	
	We formulate the lane-change problem as a nonlinear programming problem. The optimization objective is selected to minimize the acceleration and jerk of the EV in addition to the difference between the real trajectory and the goal lane.
	\begin{equation}
	C = \int_{0}^{t_f}{c_1a^2 + c_2J^2 + c_3(y - y_f) ^ 2}{d_t}
	\end{equation}
	where $t_f$ represents the total motion planning horizon (we choose $t_f = 5$ s), $a$ represents the acceleration of the ego vehicle, $J$ represents the jerk of the ego vehicle and $(y - y_f)$ represents the lateral distance from the ego vehicle to the target lane. $c_1$, $c_2$, and $c_3$ are weights. We choose $c_1 = 0.5$, $c_2 = 100$, and $c_3 = 1$ respectively.
	
	The dynamics constraints are shown in Eq. \ref{logic3}.
	\begin{equation}
	\left
	\{ 
	\begin{aligned} 
	\frac{dx}{dt} & = v\cos\theta \\
	\frac{dy}{dt} & = v\sin\theta \\
	\frac{dv}{dt} & = a \\
	\frac{d\theta}{dt} & = \omega \\
	\end{aligned} 
	\right. 
	\label{logic3}
	\end{equation}
	
	Where ($x, y)$, $v$, $\theta$, $a$, and $\omega$ represent the position coordinate, the velocity, the yaw angle, the acceleration, and the yaw rate of the ego car respectively. For simplification, in later sections, we will use $x$, $y$, $v$, and $\theta$ to denote $x_{ego}$, $y_{ego}$, $v_{ego}$, and $\theta_{ego}$. For abbreviation, $\boldsymbol{x}$ is used to denote $[x, y, v, \theta]$.
	
	The problem is expressed in Eq. \ref{optimization_problem}.
	\begin{equation}
	\begin{aligned}
	&\min\,\, C = \int_{0}^{t_f}{c_1a^2 + c_2J^2 + c_3(y - y_f) ^ 2}{d_t}\\
	&s.t.\quad
	\begin{cases}
	Eq. \ref{logic3} \\
	Eq. \ref{logic1} \\
	v_{\min} < v < v_{\max} \\
	a_{\min} < a < a_{\max} \\
	\omega_{\min} < \omega < \omega_{\max}\\
	\boldsymbol{x_0} = [x_0, y_0, v_0, \theta_0]
	\end{cases}
	\\
	\label{optimization_problem}
	\end{aligned}
	\end{equation}
	where $v_{\min}$, $v_{\max}$, $a_{\min}$, $a_{\max}$, $\omega_{\min}$ , $\omega_{\max}$ are the lower and upper bound of the ego vehicle’s velocity, acceleration and angular acceleration respectively.
	
	We separate the number of discretization points to be $N = 50$ and use trapezoidal rule to discretize the dynamic constraints. Then, we use a successive solution to solve this problem. The procedure are shown below:
	
	1)  Set $k = 0$. Select an initial trajectory ${\boldsymbol{x}^{[0]}}$.
	
	2)  At the $(k + 1)$th iteration, solve the MIQP problem and get the solution $\boldsymbol{x}^{[k]}$. 
	
	3)  Check the convergence conditions. If the convergence conditions are satisfied, go to step 4; otherwise, use the solution solved in step 3 as an initialization, set $k = k + 1$, and go to step 2.
	
	4)  Stop the successive process and obtain the solution for motion planning.
	
	\subsection{Acceleration of optimization}
	In order to guarantee the fast and exact convergence to the optimal solution, we provide a reasonable initialization of the state variables.
	
	In our method, we use a polynomial lane-change trajectory for the initialization of the state tuples ($x$, $y$, $v$, $\theta$). We adopt the cubic polynomial which has several advantageous properties including smoothness, closed-loop, continuity and 3-order derivability.
	
	The initialization can be defined in Eq. \ref{initialization}.
	\begin{equation}
	\left
	\{ 
	\begin{aligned} 
	x(t) &= a_5t^5 + a_4t^4 + a_3t^3 + a_2t^2 + a_1t + a_0 \\
	y(t) &= b_5t^5 + b_4t^4 + b_3t^3 + b_2t^2 + b_1t + b_0 \\
	\end{aligned} 
	\right. 
	\label{initialization}
	\end{equation}
	
	Considering the initial states and final states of the lane-change process, the boundary conditions are listed as follows:
	
	\begin{equation}
	\left
	\{ 
	\begin{aligned} 
	x(0) = x_0, \dot{x}(0) = v_{x, 0}, \ddot{x}(0) = a_{x, 0} \\
	y(0) = y_0, \dot{y}(0) = v_{y, 0}, \ddot{y}(0) = a_{y, 0} \\
	x(t_f) = x_f, \dot{x}(t_f) = v_{x, f}, \ddot{x}(t_f) = a_{x, f} \\
	y(t_f) = y_f, \dot{x}(t_f) = v_{x, f}, \ddot{x}(t_f) = a_{x, f} \\
	\end{aligned} 
	\right. 
	\label{initialization2}
	\end{equation}
	
	Where $x_0$, $y_0$, $v_{x,0}$, $v_{y,0}$, $a_{x,0}$, and $a_{y,0}$ represent the state of the EV before lane-change. And $x_f$, $y_f$, $v_{x,f}$, $v_{y,f}$, $a_{x,f}$, and $a_{y,f}$ represent the state of the EV at $t_f$. We call the prediction module of EV to predict the moving information of the surrounding vehicles. Then we use the predicted longitudinal position of the target vehicle at $t_f$, the width of road, the predicted velocity of the target vehicle at $t_f$, $0 m/s$, the predicted acceleration of the target vehicle at $t_f$, and $0 m/s^2$ to initialize $x_f$, $y_f$, $v_{x,f}$, $v_{y,f}$, $a_{x, f}$ and $a_{y,f}$ respectively. In Sec. \ref{sec:simulation}, we will define the movements of the surrounding vehicles.

	\section{Learning Optimal Policy}\label{sec:Learning}
	Although the complex lane-change scenario can be solved with fewer iterations using the approach described in Sec. \ref{sec:MIQP}, the method is still limited by its long optimization time, which makes it fail to be used online. An efficient approach is to combine deep learning with the optimization method. However, the trajectories output by the optimization algorithm are diverse, including both feasible path and infeasible path. And not all of the feasible paths obey traffic rules. Thus, directly learning a policy from the demonstration data is not promising. In this section, we will propose a hierarchical data-driven method called MLP-MIQP that not only determines whether to change lanes, but also outputs actions that conform to traffic rules.
	
	We define three kinds of scenarios from the trajectories given by MIQP: The ‘ill-posed’ path is a path where the EV reaches the target lane but the lateral position does not monotonically increase or a path with $|\theta| > 10deg$ at time $t_f$.
	The ‘failure’ path is a path where the EV does not reach the target lane, i.e. the MIQP could not solve the optimization problem under this scenario.	
	The ‘well-posed’ path is a path where the EV reaches the target lane with monotonically lateral position and $|\theta| < 10deg$ at time $t_f$.
	
	
	Afterward, we divide the learning process into two phases. The first phase is a ‘classification phase’, which is used to predict whether the current road condition satisfies a criterion for generating a 'well-posed' trajectory. The second phase is an ‘action generation phase’. When the first phase outputs a lane-change command, the second phase is activated and generates specific instructions for the lane-change. In addition, during the lane-change process, collision detection is continuously performed. When a possible collision is detected, it is necessary to stop the lane-change process.
	The learning algorithm framework is shown in the following:
	
	\begin{algorithm}
		\caption{Optimal Policy Learning Process}\label{euclid}
		\begin{algorithmic}[1]
			\STATE Initialize task scenario
			\STATE Classify initial states
			\IF  {initial states == ’well-posed’ or ‘ill-posed}
			\FOR	{episode p=1:M}
			\STATE {Use the output actions of MLP-MIQP}
			\STATE {Collision check}
			\IF {collision check == ‘True’}
			\STATE {conduct actions}
			\ELSIF {collision check == ‘False’}
			\STATE {take over}
			\ENDIF
			\ENDFOR
			\ENDIF
		\end{algorithmic}
		\label{algorithm1}
	\end{algorithm}
	
	\subsection{Classification Phase}
	
	To distinguish between the ‘well-posed’, ‘ill-posed’, and ‘failure’ cases, we first use a classification algorithm to judge the given road conditions. The initial states of the traffic condition (including EV and three surrounding vehicles) are chosen as input features of the classification model.
		
	
	\begin{equation}
	x_{input} = [x_{ego}, v_{ego}, x_{i}, v_{i}, a_{i}] \quad for \ i=1,2,3 
	\end{equation}
	
	We use Decision Tree (DT) with 30 learners, Naïve Bayes (NB) with Gaussian kernel, Support Vector Machine (SVM) with 3rd order polynomial kernel, K-Nearest Neighbors (KNN) with 10 neighbors, and Ensemble Classifier with 30 learners to classify these three different scenarios. 5-fold cross-validation is used to avoid overfitting.  
	
\begin{figure}[]
		\centerline{\includegraphics[height=4.00cm]{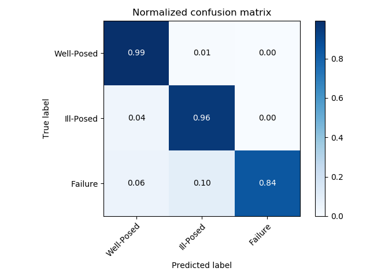}}
		\caption{Normalized Confusion Matrix for classification phase.}
		\label{Confusion_Matrix}
	\end{figure}

	We sampled 4000 trajectories generated by MIQP to train each classifier and use another 1000 trajectories to evaluate them. The accuracy of each classifier on the test set is compared in Table \ref{classifier}. Experimental results show that SVM is the most accurate model in classifying lane-change cases. The normalized confusion matrix of SVM is given in Fig. \ref{Confusion_Matrix}. In the matrix, each column represents the predicted label while each row represents the true label. It can be seen that the values in the diagonal line are approximated to 1, which indicates that the classifier is reliable.
	
	\begin{table}[]
		\centering
		\footnotesize
		\caption{Accuracy of Different Classifiers}
		\begin{tabular}{cccccc}\hline\hline \noalign{\smallskip}
			Classifiers  & DT & NB & SVM & KNN & Ensemble \\ %
			\noalign{\smallskip}\hline\noalign{\smallskip}
			Accuracy   & 86.8\%   &78.2\% &   93.1\% & 84.4\% & 91.9\% \\
			\noalign{\smallskip}\hline\hline
		\end{tabular}
		\label{classifier}
	\end{table}
	
	
	\subsection{Action Generation Phase}
	\subsubsection{Network Design}
	In the lane-change execution phase, we use a deep neural network to map the input state to a lane-change action. The time-consuming iterative optimization in traditional methods is avoided. Therefore the execution time of the proposed method is faster than the optimization-based approaches.
	
	We choose the initial states of the traffic condition (including EV and three surrounding vehicles, each feature includes the velocity, acceleration, and position) and the current states of the traffic condition during lane changing as the input features. The output features are two control signal actions $a$ and $\omega$. A multi-layer perceptron (MLP) with 10 layers, each with 10 neurons, is adopted to train this state-action policy model. The activation function is tansig function and the optimizer is Adam. 
		
	\subsubsection{Data Collection and Interaction Assumption of Surrounding Vehicles}
	There are three obstacles considered in our driving scenario, as shown in Fig. \ref{lane_change_construction}.  The initial state of these vehicles including the ego vehicle and surrounding vehicles are randomly generated. The velocity of the vehicle 1 and vehicle 2 are generated randomly between 20 to 40 km/h, following the uniform distribution. The ego vehicle is assumed to follow its proceeding vehicle in a safe following distance based on the three-second-rule, which means that the ego AVs should ideally stay at least three seconds behind its proceeding vehicle.

	Moreover, We assume that the surrounding vehicles drive in uniform motion, accelerating motion, and an intelligent driver model (IDM) \cite{treiber2000congested} respectively. For each kind of assumption, we collect 5000 trajectories, each trajectory with $5000 * (N + 1)$ data, to train our network.
	
	\section{Model Validation and Simulation} \label{sec:simulation}
	In this section, we will demonstrate the efficiency of our MLP-MIQP solver, the generality of the network, and the versatility of the MLP-MIQP. We constructed the simulation system in MATLAB 2019a and run on a laptop with 2.5Gz Intel Core i7.
	
	\subsection{Algorithm Efficiency}
	We first conduct a comparison on the efficiencies of the proposed method and other baseline methods.
	In order to simplify the computation process, the interaction between each vehicle is not considered in this experiment.
	
	A successful lane-change path planner should be able to make the AV drives to the target lane without collision with other obstacles in real-time. In our experiment, the time limitation for real-time execution is $T\leq1s$.
	
	By analyzing $5000$ sample lane-change cases with the same initial states, prediction horizon, similar objective functions and constraints in \cite{sun2017fast} for MPC (without using Dagger or logical constraints) and \cite{DWA} for DWA method. We summarize our results in Table \ref{comparison between different models}. From the results in Table \ref{comparison between different models}, it is obvious that the proposed MLP-MIQP outperforms most common used sampling based and optimization based planning module baselines.
	
\begin{figure*}[htbp]
		\centerline{\includegraphics[width=9.0in]{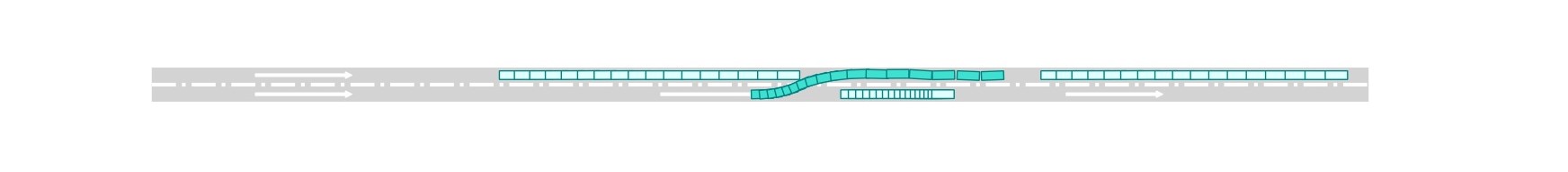}}
		\caption{Congested lane-change with proposed model}
		\label{Congested lane-change with proposed model}
	\end{figure*}

	To qualitatively show the advantages of our method, Fig. \ref{Congested lane-change with proposed model} shows an example of a congested lane-change scenario. Our method performs pretty well in such congested lane-change cases. The ego vehicle intends to change to the left lane because it needs to exit the ramp. Meanwhile, as lots of surrounding vehicles need to merge out, the gap in the target lane is not quite big enough. But the proposed path planning algorithm still finds a safe and efficient lane-change trajectory in this challenging situation.
	On the contrary, Traditional MPC without logical constraints method cannot handle this situation. The reason is mentioned in the formulation of logical constraints in Sec.{\ref{sec:MIQP}}.
	Furthermore, compared with DWA method, we found that they will tend to become less efficient in our scenario due to the high speed of obstacle vehicles.
	
	\begin{table}[]
		\centering
		\footnotesize
		\caption{Comparison Between Different Model}
		\begin{tabular}{ccccc}\hline\hline \noalign{\smallskip}
			&  Success rate& Computation time &  &  \\ %
			\noalign{\smallskip}\hline\noalign{\smallskip}
			MLP-MIQP &85.171\% &6ms     \\
			MPC\cite{sun2017fast}  &65.905\%   &230ms    \\
			DWA\cite{DWA}  &57.897\%   &489ms    \\
			\noalign{\smallskip}\hline\hline
		\end{tabular}
		\label{comparison between different models}
	\end{table}
	
	\subsection{Generalization Capability}
%
	As stated in Sec. \ref{sec:Learning}, we only use the 'well-posed' data to train the MLP. However, when we use the MLP-MIQP model to generate a trajectory that based on an 'ill-posed' scenario, the proposed algorithm can still generate a path that meets the condition of a 'well-posed' trajectory.
	
\begin{figure*}[htbp]
		\centering
		
		\subfigure[Ill-posed trajectories planned by MIQP]{
			\begin{minipage}[t]{1.00\linewidth}
				\centering
				\includegraphics[width=7.0in]{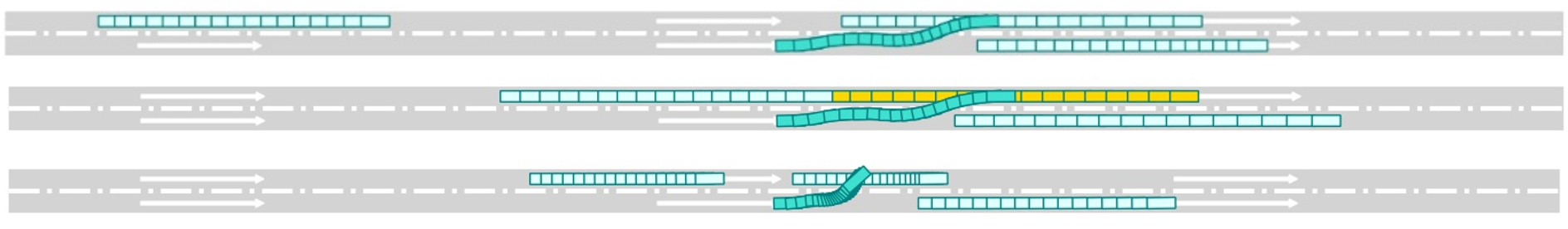}
				\label{a}
			\end{minipage}%
		}%
		
		\subfigure[After calling the MLP-MIQP, three listed scenarios are rectified]{
			\begin{minipage}[t]{1.00\linewidth}
				\centering
				\includegraphics[width=7.0in]{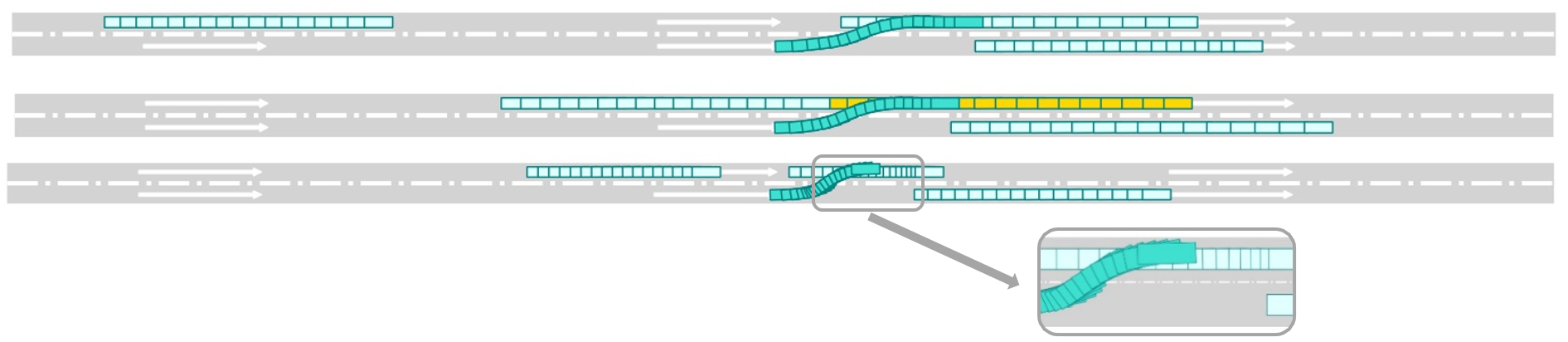}
				\label{b}
			\end{minipage}%
		}%
		\centering
		\caption{\label{Generality of MLP-MIQP}Generality of MLP-MIQP. 
		}
	\end{figure*}

	Fig. \ref{Generality of MLP-MIQP} shows 3 driving scenarios where MIQP returns an ill-posed trajectory in comparison to the results solved by MLP-MIQP. Fig. \ref{a} shows the result of the MIQP model and Fig. \ref{b} shows the result of the MLP-MIQP model under the driving conditions.
	
	

	We test 1000 ill-posed scenarios in total. From the results, we found that the proposed model outputs well-posed trajectories on	$22\%$ scenarios. It indicates that the neural networks can solve a broader range of lane-changing problems than optimization-based model MIQP. Moreover, it also indicates that the proposed MIQP-MLP can be well generalized to the environment which is different from the one it is trained with.
	
	Remark I: To explain the generalization capability, we first consider the reason for why ‘ill-posed’ routes are planned. In the view of the MIQP algorithm, such an ‘ill-posed’ route is a solution that makes the lowest cost while satisfying the constraints. However, since the ‘action generation phase’ only retains good planning results, it learns the ‘well-posed’ paths that the vehicle can drive. Thus, for such ‘ill-posed’ cases, it achieves the similar planning results as ‘well-posed’ trajectories. Although we can avoid the generation of bad paths by formulating much more complex constraints, the higher nonlinearity would make MIQP become more difficult to converge. Further, it is more difficult for neural networks to learn such a complex optimization algorithm.
	
	\subsection{Versatility}
	
	In many literature, e.g. \cite{MIQP,ji2016path}, the driving policies of obstacle vehicles are assumed to be constant velocity or constant accelerating. These assumptions ignore interactions between vehicles. While other complex policies considering the interaction between vehicles are always more realistic but computationally expensive. On the contrary, our proposed method leverages the on-line output of deep learning to tackle with off-line time-consuming problems.
	
	Thus, we assume constantly moving, uniformly accelerating, and the intelligent driver model (IDM) separately to further compare the lane-change success rate for the different assumptions of obstacles driving polices.  $5000$ driving scenarios are sampled for each assumption. Follow the definition of the 'well-posed' lane-change trajectories, the success rate under each driving policy is shown in table \ref{different assumption}. Even for a sophisticated policy like Intelligent Driver Model (IDM), the MLP-MIQP enables handling the interactive scenarios via online computing.
	
	\begin{table}[]
		\centering
		\footnotesize
		\caption{lane-change Success Rate Under Different Driving Policy}
		\begin{tabular}{lc}\hline\hline \noalign{\smallskip}
			Driving behavior of surrounding vehicles  & Success rate  \\ %
			\noalign{\smallskip}\hline\noalign{\smallskip}
			Uniformed speed &88.75\%   \\
			Uniformed acceleration  &85.17\%     \\
			Intelligent Driver Model  &91.95\%     \\
			\noalign{\smallskip}\hline\hline
		\end{tabular}
		\label{different assumption}
	\end{table}
	
	Note that the success rate of lane-change under the assumption of IDM is even higher than the other two assumptions. This is due to the interaction between vehicles so the success rate is high. This indicates that our method is worthy of further exploration by using a more realistic and interactive traffic flow model.
	
	\section{Conclusion and Future Work}
	In this paper, we devise a fast MLP-MIQP algorithm for solving more generalized and complex AVs motion planning problem. We introduce the logical constraints to tackle obstacle avoidance in a congested lane-change scenario. Meanwhile, a hierarchical machine-learning framework is proposed to improve the solution space, learning speed, and lane change performance. In order to improve the learning efficiency and effectiveness, a SVM based traffic state classification layer is used to select the training samples. 
	
	From the experimental results, we find that our model exhibits strong generalization capability. Additionally, we show that our model can maintain a high lane-change success rate on cases with complex vehicle-to-vehicle interactions. In the future, we will try some efficient technique, e.g. DAgger (Dataset Aggregation) to improve our current imitation framework and explore the usage of other algorithms including deep reinforcement learning, and evolutionary strategies on AV motion planning.
	
	\bibliographystyle{IEEEtran}
	\bibliography{./bib/mylib} 

\end{document}